\documentclass[10pt,twocolumn,letterpaper]{article}

\usepackage{wacv}
\usepackage{times}
\usepackage{epsfig}
\usepackage{graphicx}
\usepackage{amsmath}
\usepackage{amssymb}
\usepackage{booktabs}
\usepackage{subfig}
\usepackage{makecell}
\usepackage{multirow}
\usepackage{adjustbox}
\usepackage{draftwatermark}

\def\wacvPaperID{1039} 

\wacvalgorithmstrack   

\wacvfinalcopy 

\ifwacvfinal
\usepackage[breaklinks=true,bookmarks=false]{hyperref}
\else
\usepackage[pagebackref=true,breaklinks=true,colorlinks,bookmarks=false]{hyperref}
\fi

\SetWatermarkText{DRAFT Version}
\SetWatermarkScale{.5}
\begin{document}

\title{\huge AttTrack: Online Deep Attention Transfer for Multi-object Tracking}

\author{Keivan Nalaie\\
McMaster University\\
{\tt\small nalaiek@mcmaster.ca}
\and
Rong Zheng\\
McMaster University\\
{\tt\small rzheng@mcmaster.ca}
}

\maketitle
\thispagestyle{empty}

\begin{abstract}
Multi-object tracking (MOT) is a vital component of intelligent video analytics applications such as surveillance and autonomous driving. The time and storage complexity required to execute deep learning models for visual object tracking hinder their adoption on embedded devices with limited compute power. In this paper, we aim to accelerate MOT by transferring the knowledge from high-level features of a complex network (teacher) to a lightweight network (student) at both training and inference time. The proposed AttTrack framework has three key components: 1) cross-model feature learning to align intermediate representations from the teacher and student models, 2) interleaving the execution of the two models at inference time, and 3) incorporating the updated predictions from the teacher model as prior knowledge to assist the student model.  
Experiments on pedestrian tracking tasks are conducted on the MOT17 and MOT15 datasets using two different object detection backbones YOLOv5 and DLA34 show that AttTrack can significantly improve student model tracking performance while sacrificing only minor degradation of tracking speed. 
\end{abstract}

\section{Introduction}
Multi-object tracking (MOT) is a fundamental problem in computer vision that aims to identify trajectories in video frames. MOT is a key building block of many applications such as human-computer interaction, surveillance, and autonomous driving tasks. Existing methods handle a MOT task by treating it as two sub-problems: \textit{object detection} and \textit{object association}. Object detection identifies bounding boxes of objects of interest in each frame whereas object association links detected objects to form trajectories over time.  

As neural network models for MOT become more complex, improved accuracy usually comes at the cost of longer inference time. 
To accelerate the execution of deep models, techniques such as model quantization~\cite{rastegari2016xnor,hubara2017quantized,cai2017deep} and pruning~\cite{li2016pruning,molchanov2016pruning,zhou2019deconstructing} have been widely utilized to reduce computations and redundant connections. Computation acceleration from model quantization is generally hardware dependent~\cite{gholami2021survey}. Extensive parameter tuning is required for network pruning to work without significant loss of accuracy~\cite{cheng2018model}. Recently, a few works utilized temporal\cite{chin2019adascale,meng2020arnet,liu2020continuous} and spatial~\cite{xu2018deepcache} correlations by finding configurations (e.g., frame rate, frame resolution) that achieve a good trade-off between computation complexity and model performance. Unfortunately, the optimal configuration is not only input sensitive but also dependent on run time environments such as the available CPU or memory resources.

Another line of work to reduce model complexity is through knowledge distillation (KD)~\cite{zagoruyko2016paying}. KD uses soft labels generated by a large model (teacher) to train a small neural network with fewer parameters (student). The soft labels provide useful information that allows the student model to learn the behavior of the teacher to improve generalization. However, the lack of distilled knowledge for the student model during inference may hinder it from correctly detecting harder instances (e.g., crowded scenes with smaller objects). 

In this work, we propose an \textit{ attention transfer approach for object tracking} aiming to exploit the knowledge of a teacher model at both training and inference stages. The proposed online deep attention transfer network (AttTrack), is inspired by the idea of attention transfer first proposed in~\cite{zagoruyko2016paying}. Unlike existing tracking methods, our detection model receives additional information in the form of previously detected objects from the teacher model. We only run the teacher model every few frames (called \textit{key frame}) during inference to improve the representation capability of the student model and help it to discover likely object positions in the remaining frames. The student model can gain information about obstructed or barely visible objects by leveraging extracted information from outputs of the teacher model on prior frames at no extra cost. The student model is trained to fuse extracted features of input frames and the detection estimation from the teacher model.

Extensive experiments show that AttTrack improves the tracking performance of small models with marginal increases at interference time. Choosing the intervals to run the teaching model, results in different trade-offs between performance and efficiency. In summary, the main contributions of this paper are as follows:
\begin{itemize}
\item We conduct an empirical study to investigate the impacts of model size on tracking performance and speed.
\item We propose an end-to-end trainable AttTrack framework to transfer the knowledge of a complex teacher model to a lightweight student model at both training and inference time.  
\item Extensive testing on the MOT17 and MOT15 datasets demonstrates the effectiveness of AttTrack. For example, our approach can improve the MOTA score on YOLOv5 and DLA34 architectures by up to 12\% with comparable computation time when compared to existing attention-based methods. 
\end{itemize}
The remainder of the paper is structured as follows. We first describe related work in Section \ref{related_work_section} and study the impact of model size in tracking performance and speed in Section \ref{motivation}. In Section \ref{framework} details of AttTrack is presented. We present experimental results in Section \ref{experiments}, followed by a conclusion in Section \ref{conclusion}. 

\section{Related Work} \label{related_work_section}
\subsection{Knowledge distillation} Knowledge Distillation (KD) was first proposed by Hinton \textit{et al}.~\cite{hinton2015distilling} , which aims to train a student (a smaller and faster) model by transferring knowledge from a teacher (a bigger and slower) model.
This knowledge in the form of softened outputs of the teacher model is more informative than one-hot vectors in training data. Subsequent studies improved upon~\cite{hinton2015distilling} and devised various ways to ease the training of small models with few trainable parameters. Romero \textit{et al}.~\cite{romero2014fitnets} proposed FitNet, which uses the intermediate representation learned by a teacher model to change the structure of a small model from being wide and shallow to narrow and deep. Knowledge transmission was considered as a distribution matching problem in~\cite{huang2017like}. Despite success of KD in classification problems, the needs for bounding box regression and heatmaps estimation in object detection introduce additional obstacles. To extend KD to object detection Chen \textit{et al.}~\cite{chen2017learning} included a feature imitation loss into the detection loss to use the intermediate features of the teacher as hints for the student model. In~\cite{li2017mimicking}, the authors devised MIMIC, an extended KD for detector models by employing a fully convolutional feature mimic architecture to transfer knowledge for each pixel individually. In order to avoid teacher supervision for background regions, Mehta \textit{et al.}~\cite{mehta2018object} introduced \textit{objectness scaled distillation} for one-shot object detectors. Similarly, our method uses the attention mechanism of the teacher to train the student model on softer labels. However, the key distinction between our AttTrack and the aforementioned approaches is that during inference time, the student model uses real teacher outputs, to calibrate tracking outputs and achieve better accuracy.

\subsection{Attention mechanisms in object tracking}There is a long line of studies that combine the concept of attention with machine learning. Human attention mechanism theories~\cite{rensink2000dynamic} inspired early efforts on attention-based learning such as \cite{larochelle2010learning,denil2012learning}.
Attention has been used in a wide range of machine learning tasks including deep learning-based video object tracking. Fiaz \textit{et al}. \cite{fiaz2020improving} proposed a channel attention method that gives higher weights to channels that help with target classification and localization.  Huang \textit{et al}.~\cite{huang2021siamatl} proposed an attentional online update paradigm for siamese visual tracking to improve the performance of a tracker by utilizing knowledge extracted from prior tracking tasks. In ~\cite{thanikasalam2019target} residual attention modules are introduced in similarity tracking at multiple levels of feature representation, resulting in improved discrimination quality for similarity searching. Zhang \textit{et al}.~\cite{zhang2021toward} created an attention retrieval network that uses learning masks to conduct soft spatial constrains on features from a  tracking backbone network, mitigating the impact of background clutter. 

\subsection{Trainable attention mechanism for object detection} Researchers have explored attention mechanisms in object detection to enhance feature representation. The encoder and decoder stages of the object detecting system presented in~\cite{dai2021dynamic} use a dynamic attention approach. The attentions are determined by size, feature dimension, and spatial features using a convolution-based encoder. In ~\cite{ying2019multi} the authors proposed a feature pyramid network to object detection in remote sensing images, adapting two types of attention mechanisms: a) global spatial attention that extracts spatial location-related features to improve the positioning ability of the object detector, and b) pixel feature attention that expands the size of receptive fields that makes the model learn more image details. Reverse attention was explored in ~\cite{chen2018reverse} to assist top-down side-output residual learning in order to acquire more accurate residual features and handle missing object areas and details. In~\cite{wang2019salient}, Wang \textit{et al}. applied a pyramid attention module in their deep saliency model to give more weight to salient regions while extracting multi-scale characteristics from input images. In contrast to the previous researches, to achieve domain sensitivity in object detection, Wang \textit{et al}. \cite{wang2019towards} utilized a domain attention module for universal object detection. 

AttTrack is orthogonal to the previous attempts and can be used in conjunction with other attention-based methods. It takes advantage of the knowledge of more complex teacher models at both training and inference time.


\section{Preliminary Study of Model Sizes on Tracking Performance}
\label{motivation}
To motivate our approach we first conduct empirical evaluations on the MOT17 dataset to assess the performance of tracking models with different model sizes.
\subsection{Computation time break-down}
Object detection, feature extraction, and object association are the three components of a conventional deep neural network (DNN)-based tracker. In ~\cite{nalaiedeepscale}, it has been reported that object detection is the most time-consuming part of the tracking process. Therefore, this study focuses on reducing the computing cost of object detection while maintaining the overall tracking performance.
\subsection{Experimental Setting}
\label{experimental_settings}
\textbf{Object detection.}
There are two main categories of approaches in DNN-based object detection. \textit{Two-stage approaches} first extract regions of interest (RoIs) and then classify and regress the RoIs. R-CNN~\cite{girshick2014rich}  and Faster-RCNN~\cite{fasterrcnn} are two widely used object detection models in this category. In the second category, \textit{one-stage approaches},  directly identify and regress objects of interest. For example, YOLO~\cite{redmon2016you} divides an input image into $ S \times S$ grids and performs region classification and regression.

In this work, we choose FairMOT~\cite{zhang2020fairmot}, a state-of-art (SOTA) one-stage object detection model for three reasons. First, one-stage object detectors tend to be faster than two-stage object detectors. Second, with the YOLOv5~\cite{jocher2021ultralytics} backbone, FairMOT results in a good trade-off between speed and computation complexity. Third, for each identified object, FairMOT computes re-ID features, which can be utilized in object association and tracking. 

\textbf{Model size.}
We evaluate three models following the YOLOv5 architecture but with different sizes: YOLOv5, YOLOv5-mid (a model with half of the channels in each layer of the base model), and YOLOv5-small (a model with a quarter of the number of channels in each layer of the base model). All three models are pre-trained on the COCO dataset (for object detection task) and then trained on the MOT-17 dataset (for multi-object tracking task).

\textbf{Performance metric.}
MOTA and IDF1(F1) are commonly used in MOT to assess tracking performance. An MOTA score is calculated as follows:
\begin{equation}
    MOTA = 1 - \frac{\sum_{t} FP_t + FN_t + IDSW_t}{\sum_{t} GT_t},
\end{equation}
where $t$ denotes the frame index, $FP_t$, $FN_t$, and $IDSW_t$ denote the number of false positive, false negative, and ID switched objects, respectively, and $GT$ denotes the number of ground truth bounding boxes. 

Experiments are conducted on an NVIDIA GTX 3080 graphical card with 8 GB GDDR6, running a docker on Ubuntu 20.04. The system is built using Pytorch v1.8 and CUDA v11.3. 

\subsection{Results and Observations}
The tracking performance of the three models on the validation dataset is shown in Table \ref{table:models_and_perfromances_init}. It is clear that lowering model size leads to a decrease in tracking accuracy but accelerated inference (measured in frame per second (FPS)). YOLOv5-mid, for example, is faster than the base model at the cost of a 5.5\% drop in MOTA score. Similarly, YOLOv5-small suffers around a 23.5\% drop in MOTA but is 1.35 times faster than the base model. 

\begin{figure*}[]
\centering
\subfloat[Extracted trajectories for MOT17-04 using YOLOv5-small model ]{\includegraphics[scale=0.18]{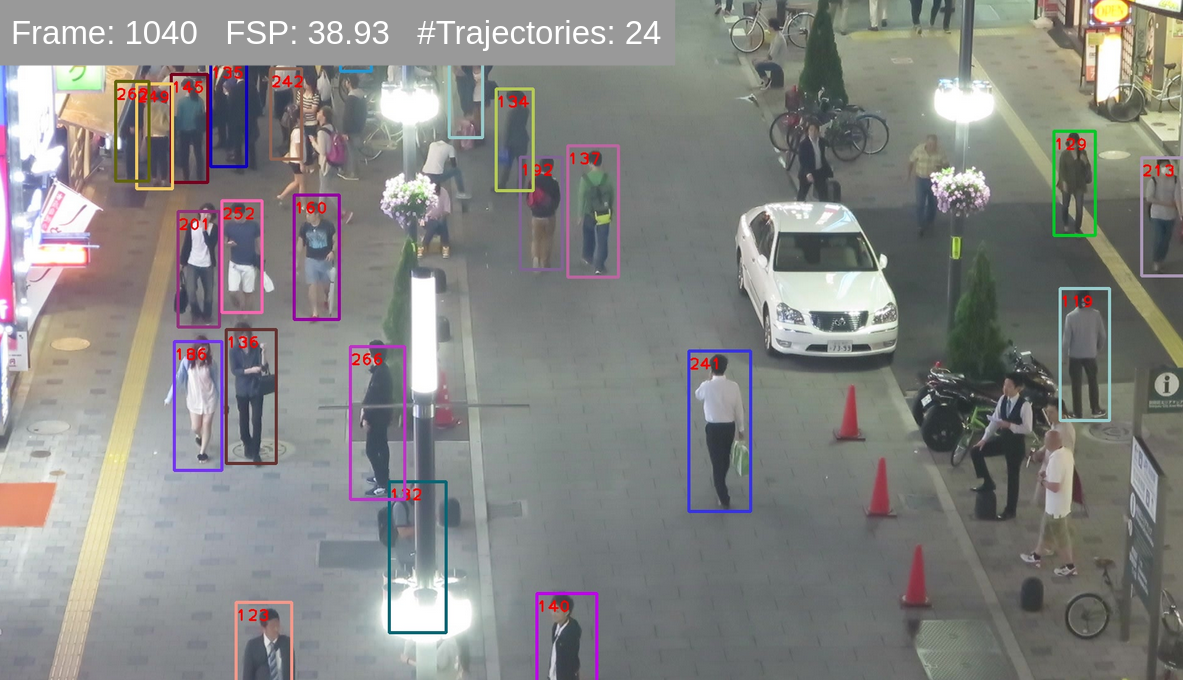}}
\subfloat[Extracted trajectories for MOT17-04 using YOLOv5 model]{\includegraphics[scale=0.18]{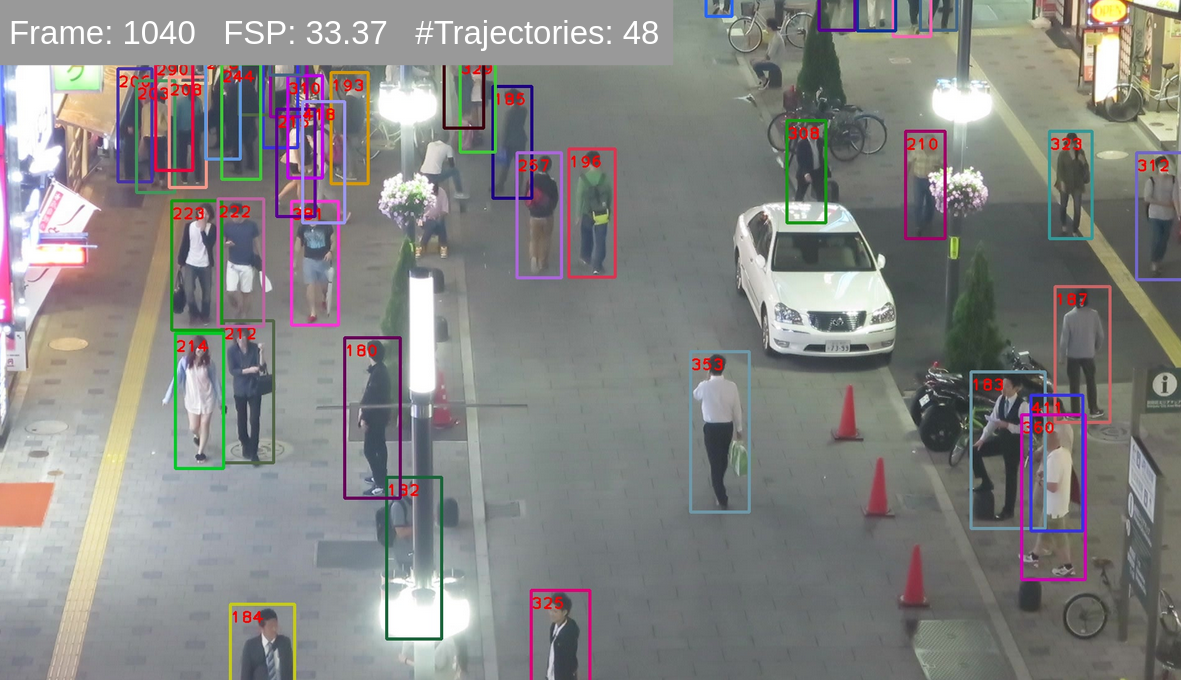}}%
\caption{Demonstration of tracking results from a small and a large model. The YOLOv5-small model performs less accurately than the YOLOv5 model due to partially occluded or small-size objects.}
\label{fig:tracking_visualization}
\end{figure*}

Figure \ref{fig:tracking_visualization} illustrates the tracking performance for the full model and the small model. Compared to the full model, the small model fails to detect some objects, especially those that are far away, partially occluded, or small sizes. Therefore, our main goal is to train an efficient small neural network using knowledge from a big model to attain comparable performance. Unlike existing work on attention-based approaches, knowledge transfer is performed both during the training and inference stages.  
\begin{table}[h]
\footnotesize
\centering
\caption{Impact of model size on tracking performance}
 \begin{tabular}{c|c|c|c|c}
 \hline
  Model&IDF1(\%)$\uparrow$&MOTA(\%)$\uparrow$&FPS$\uparrow$&Parameter size\\
 \hline
 YOLOv5&65.90&62.40&43.93&5.01 M\\
 YOLOv5-mid&63.20&56.90&46.16&1.38 M\\
 YOLOv5-small&44.70&38.90&59.32& 0.31 M\\
 \hline
\end{tabular}
\label{table:models_and_perfromances_init}
\end{table}




\section{The AttTrack Framework}
\label{framework}
\begin{figure}[t]
\centering
\subfloat[Teacher/student detection performance comparison: a powerful teacher model vs a small student model ]{\includegraphics[scale=0.20]{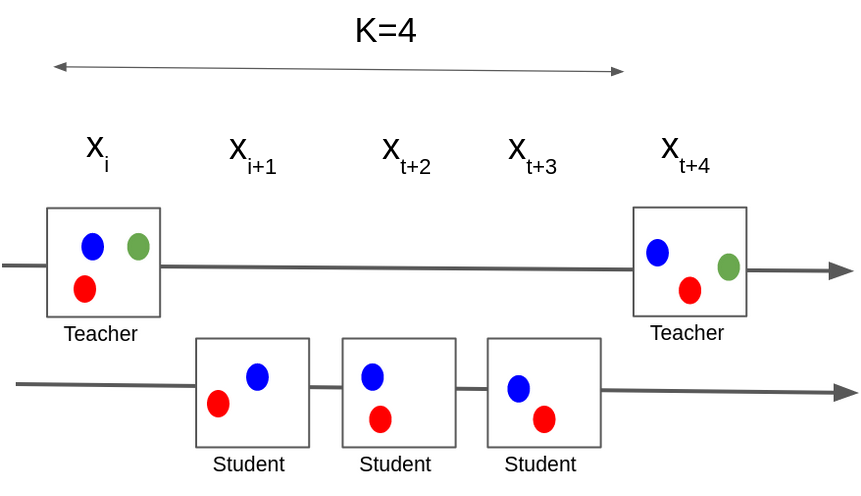}}\\
\subfloat[Inference stage: attention transfer for two cycles]{\includegraphics[scale=0.20]{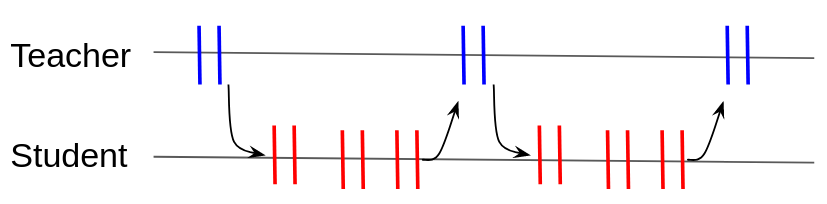}}%
\caption{Schematic illustration of attention transfer. The teacher model is used every $K=4$ frames.}
\label{fig:t_s_schematic}
\end{figure}
\begin{figure*}
\centering
\includegraphics[scale=0.30]{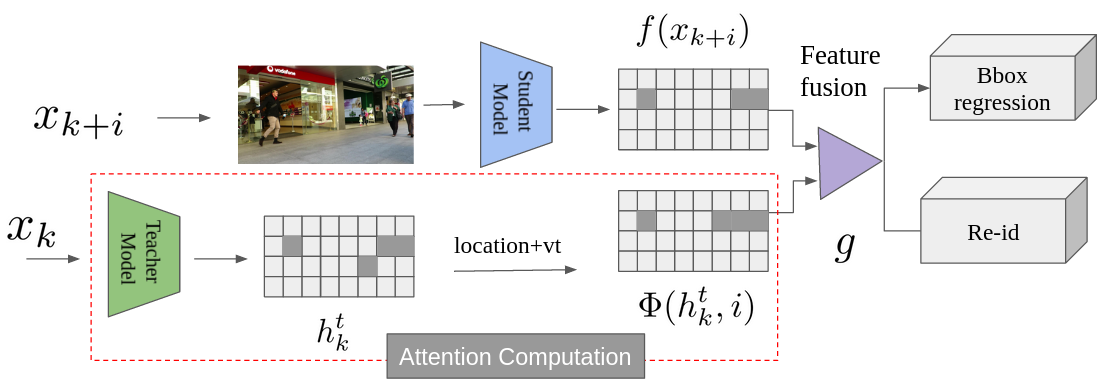}
\caption{System architecture: AttTrack applies a teacher model every $K$ frames, and computes updated states $\Phi(h_k^t, i)$ (attention) for intermediate frames based on the teacher's predicted heatmap $h_k^t$ for frame $k$. The updated state is then fused with the prediction on frame $k+i$ by a student model using a fusion network, resulting in object bounding boxes and re-identification features for frame $k+i$.}
\label{fig:system_arch}
\end{figure*}

Figure \ref{fig:t_s_schematic} shows the schematics of the proposed online attention transfer approach. AttTrack employs a teacher model to accurately detect objects from every $K$ frames at the inference time, and a student model combines this knowledge in its tracking model in the interim $K-1$ frames as depicted in Figure \ref{fig:t_s_schematic}.

We formulate the video based object detection and tracking problem as follows: given a set of $N$ input frames $X=\{x_1,x_2,...,x_N\}$ where $x_n\in \mathcal{R}^{3\times H\times W}$, the objective is to first obtain set of $M_i$ bounding boxes $B_i=\{b_{i,1},b_{i,2},...,b_{i,M_i}\}$, where $b_{i,j}=\{rect_{i,j},\phi_{i,j}\}$, $rect_{i,j}$ denotes the 4-dim vector (center coordinates, height and width) associated with the $j$th bounding box and $\phi_{i,j}$ represents the extracted visual features of the bounding box in frame $i$. Consequently, object tracking aims to construct the set of trajectories $T_{t}=\{\mathcal{B}_t,id_t\}$, where $\mathcal{B}_t$ is the set of detected bounding boxes in trajectory $t$, $id_t$ denotes the trajectory ID.
\subsection{Online Attention Transfer}
Modern object detectors such as FairMOT output heatmaps in addition to bounding boxes, where the value of each pixel in the heatmap is its likelihood of being an object center.   
Let the heatmap output by the teacher for keyframe $k$ be $h^t_k$:
\begin{equation}
    h^t_k = \mathcal{H}_t(x_k)
\end{equation}
where $\mathcal{H}$ represents the function associated with the heatmap head of the teacher model. Then, we denote the student model output $y_{k+i}^s$ at frame $k+i$ as below:
\begin{equation}
    y_{k+i}^s = g(f(x_{k+i}),\Phi(h_k^t,i))
\end{equation}
where $i$ stands for number of frames after the keyframe $k$ and $\Phi(h_k^t,i)$ extrapolates the the heatmap of teacher in frame $k$ to get its heatmap in frame $k+i$, $f$ is the generated features by backbone of the student model and $g$ is the fusion function to be explained in Section \ref{section_network_design}.

Figure \ref{fig:system_arch} depicts system architecture of AttTrack. A non-key frame is processed by the student model, to generate intermediate features. The student model then incorporates updated attention features based on the heatmap of teacher on the most recent keyframe. With the fused features, bounding box regression and re-Id networks are applied to generate the bounding box and re-ID features of each object. During tracking, a trajectory is constructed from detected objects that are similar in appearance-based re-ID features and have a large intersection of union (IoU). Specifically, object association is done in two steps: first, visual features are used to match a trajectory and a detected object. Second, if a match is found, the IoU measure is applied to determine whether a true match is obtained. If the object is matched to a trajectory, the trajectory is extended; otherwise, a new trajectory is initiated. Cosine similarity is used in computing the similarity of visual-based features. A Kalman filter \cite{zhang2020fairmot} is then applied to update the position state of each trajectory in the current frame.

Since the teacher model is applied for frames between two keyframes, the heatmaps (attention) of teacher are outdated for any frame in-between. To extrapolate the heatmap of teacher for these frames, we devise an attention update approach next. 

\subsection{Attention Update}
\label{sec_attention_update}
\begin{figure}[t]
\centering
\subfloat[ States of outputs of teacher are updated using linear kinematic equation]{\includegraphics[scale=0.20]{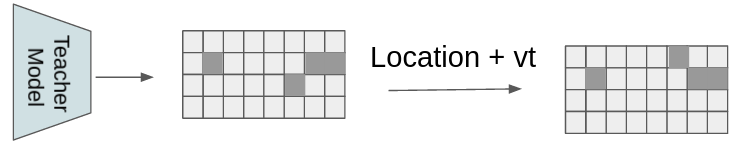}}\\
\subfloat[Inference stage: attention transfer]{\includegraphics[scale=0.20]{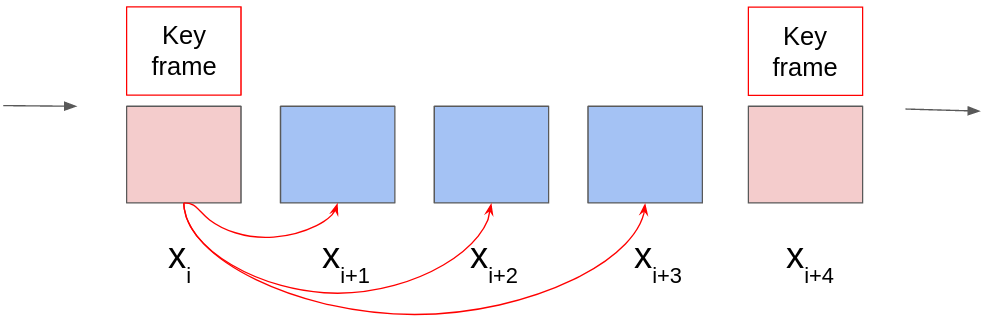}}%
\caption{Attention state update. We choose $K=4$ as number of frames between every keyframes. At frame $F_k$ and $F_{k+4}$ teacher model is performed whereas between these two keyframes the student model is applied.}
\label{fig:attention_update}
\end{figure}

The knowledge computed in earlier frames by the teacher is beneficial to the student model. However, due to the presence of moving objects, such information becomes more outdated as the time elapses between the current frame and the most recent key frame. Therefore, updates need to be made from the teacher heatmap (Figure \ref{fig:attention_update}). Consider $B^t_k$ the set of objects detected by the teacher model in frame $k$. We first estimate the velocity of each identified object based on bounding box locations from previous frames. The velocity is then used to predict the subsequent locations of the corresponding object in frame $k+i, i =1, 2,..., K-1$ using a simple linear kinematic equation. The heatmaps are updated accordingly.

The updated heatmaps are most beneficial when the student model fails to detect an object due to poor visibility. However, when object movements are irregular, a new object enters the scene or an object exits the scene, the information of teacher can still be stale. Therefore, the updated heatmaps should be combined with the prediction from the student model for the current frame.

\subsection{Network Design}
\label{section_network_design}
\begin{figure}
\centering
{\includegraphics[scale=0.23]{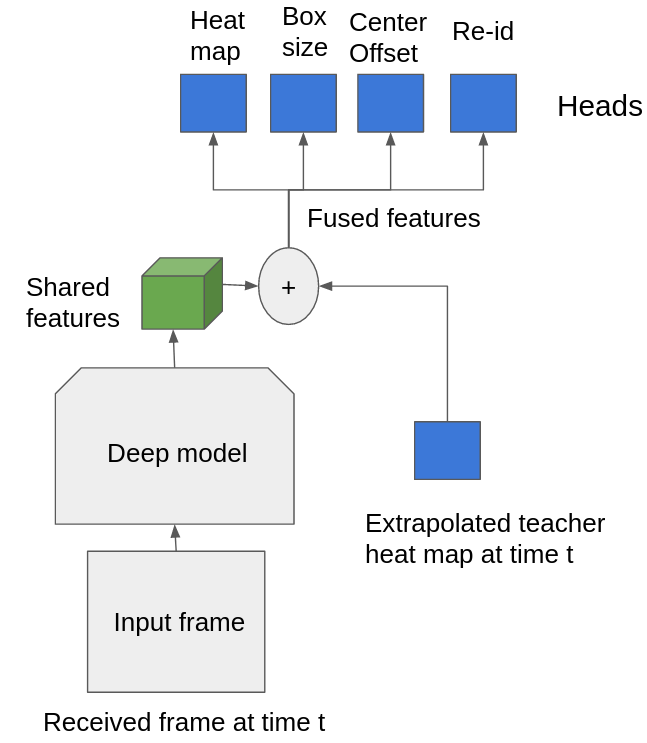}}%
\caption{Student network architecture. The student model has fewer parameters and takes two inputs: the input-frame at time $t$ and estimated heatmaps of teacher up to time $t$.  }
\label{fig:network_design}
\end{figure}
For the teacher model, we can utilize any existing object detection backbone such as DLA34~\cite{yu2018deep}. The student model, like the teacher model, creates bounding boxes and re-ID features for observed objects.
For faster computations, the student model employs fewer parameters than the teacher model in its network backbone. The student model receives attention features and input image as inputs and fuses the attention features and its own calculated features as: 
\begin{equation}
    \label{eq.fusion}
    g(f(x_{k+i}),\Phi(h_k^t,i)) = (f(x_{k+i}),\Phi(h_k^t,i))
\end{equation}
where fusion function $g$ appends the extrapolated features from teacher with new features calculated by the student backbone. The fused features are fed into heatmap and re-ID branches as defined in~\cite{zhang2020fairmot}. 

The student model is trained using the following loss function:
\begin{equation}
\label{eq:studnt_loss}
\begin{split}
    &\mathcal{L}_{student}  = \\ &\frac{1}{2}(\frac{1}{e^{\omega_1}}(\mathcal{L}_{heatmap}+\mathcal{L}_{box})+\frac{1}{e^{\omega_2}}\mathcal{L}_{identity}+\omega_1+\omega_2),\\
    \end{split}
    \end{equation}
{which consists of learnable task based parameters $\omega_1$ and $\omega_2$, heatmap loss, box-size loss, and re-identification loss defined in~\cite{zhang2020fairmot}}. 
\subsection{Cross-model Feature Learning}
\label{sec:cross_model_feature_learning}
\begin{figure}[t]
\centering
\includegraphics[scale=0.18]{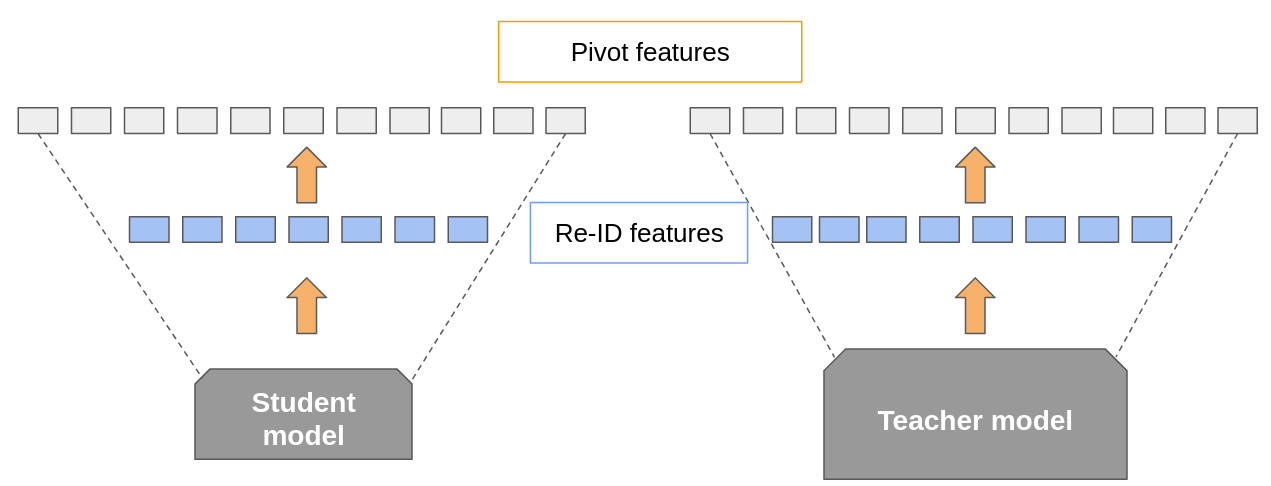}
\caption{Cross model feature learning. In the object association step, pivot features for EFM and re-ID features for IFM are used. The re-ID features of the student model in IFM are aligned with the teacher model. }
\label{fig:crossmodel}
\end{figure}
Switching from one model (for example, teacher) to another (for example, student) can result in re-ID features that follow different distributions. Running AttTrack on a video clip necessitates multiple transitions between the teacher and student models. When re-ID features mismatch between teacher and predictions of student , identity fragmentation occurs, leading to reduced tracking accuracy. To mitigate the domain gap between the re-ID features produced by the student and the teacher models, we propose two cross-model feature learning approaches.

\textit{Explicit Feature Mapping (EFM):} We use pivot features to induce correlation between the computed re-ID features by the teacher and the student models. This is done by applying a single linear layer to map the re-ID features to the number of identities (encoded as a one-hot vector) in the training set. Both student and teacher models are subjected to the linear layer. By mapping each identity to those learned pivots in the training time, this approach lowers the distance across two model domains in the inference time. 

\textit{Implicit Feature Mapping(IFM):} the former method requires an additional compute unit in both the student and teacher models, resulting in increased total computation costs. In the second approach, we perform feature mapping implicitly for the student model by mincing the re-ID feature layer in the teacher model. During training, the extracted features from the teacher model are used as an additional supervision signal and the loss function is updated as:
\begin{equation}
    \mathcal{L}_{W} = \mathcal{L}_{student}(W) + \mathcal{L}_{id-att}(W).
\end{equation}
where $\mathcal{L}_{student}$ is the loss function in Eq.\ref{eq:studnt_loss} and $\mathcal{L}_{id-att}$ is L2 loss of re-ID features.

\section{Performance Evaluation}
\label{experiments}
To qualify the performance of AttTrack, we conduct experiments on a pedestrian tracking task. The MOTA score is used to evaluate tracking accuracy, while FPS is used to quantify tracking speed.   

\subsection{Datasets}
We evaluate AttTrack on pedestrian tracking tasks on two MOT datasets. We use 11 training video clips in the publicly accessible dataset MOT17. We also provide the results of AttTrack on MOT15 \cite{motchallengeweb}, which in addition to low-resolution street view videos it includes  videos from PETS~\cite{ellis2010pets2010} and KITTI~\cite{geiger2012we} datasets. Since the ground truth of test sequences is not made public, each video is split into two halves with the first half in the training set and the second half in the test set. We also use the COCO dataset to pre-train the models. To evaluate the tracking performance of our models, we utilize the official evaluation method from MOT Challenge~\cite{motchallengeweb}. 

\subsection{Implementation Details}
Experiments are conducted using the same hardware and software setup as in Section \ref{experimental_settings}. We use the Adam optimizer to train our model across 35 epochs, with a starting learning rate of 1e-5 that lowers every 25 epochs. A batch size of 12 is used. Rotation and scaling are applied to augment the training set. The input frame size is $1088\times 608$ pixels.  

To evaluate the performance of AttTrack, we consider two backbone models for the student model. The first is DLA34-small, which offers a good trade-off in tracking performance and speed, and is based on DLA34 used in \cite{zhang2020fairmot}. It has in total 16.55M parameters. The second one, YOLOv5-small, has the same architecture as YOLOv5~\cite{jocher2021ultralytics} but with only one-fourth of the parameters, allowing for fast computations and acceptable performance. 


\subsection{Baseline Methods}
To evaluate AttTrack, we consider the following baselines: \textit{Teacher-only} and \textit{Student-only}: object detection in the tracking pipeline only uses teacher and student models, respectively. \textit{Naive-Mix}: which alternates between a teacher and a student model every $K$ frames and merge the tracking outputs of the two with no further information sharing in training or inference; AttTrack w/o attention update (\textit{AttTrack-no-update}), which transfers attention from Teacher to Student at inference but does not update the attention (i.e., $\Phi(h_k^t, i) = h_k^t$); and \textit{Layerwise}, which is similar to Naive-Mix but allows layerwise attention transfer from the teacher to the student model in training\cite{zagoruyko2016paying}.

\subsection{Online-based Attention Transfer}

Table \ref{table:fullmodels} summarizes the performance of teacher models with full-fledged DLA34 and YOLOv5 backbones. The results for the student models with and without AttTrack, AttTrack w/o attention update are given in Table \ref{table:attention_or_not}.  In Table \ref{table:attention_or_not}, with or without AttTrack, the teacher model is executed every 6 frames. The difference between the two lies in whether attention transfer is performed or not. Similar to the results in Section \ref{motivation}, smaller models have fast inference time but suffer from lower accuracy. DLA34-small without AttTrack, for example, is $1.52\times$ faster at the price of 6.9 percent MOTA degradation. AttTrack improves the tracking performance of the student model by 1.6\% and 5\% for DLA and YOLOv5, respectively. This shows the effectiveness of attention transfer from the teacher model. Because AttTrack invokes the teacher model every 6 frames, the running time of AttTrack is longer than those without. In Table \ref{table:attention_or_not}, we further compare AttTrack and AttTrack-no-update. As expected, AttTrack-no-update is faster due to less computation but has slightly degraded performance. The relative small gap between the two can attributed to small changes in the scenes when $K = 6$.  

To better understand the impact of $K$ on AttTrack, Table \ref{table:fixed_w_size} lists the results of different student models under various $K$s. As expected a smaller $K$ means more frequent execution of the teacher model, resulting in slower processing time and more accurate tracking outputs, and vice versa. YOLOv5-small runs faster than its DLA34 counterpart but with lower accuracy.

\begin{table}[h]
\footnotesize
\centering
\caption{Performance of Teacher-only and Student-only baselines on the MOT17 dataset}
 \begin{tabular}{c|c|c|c}
 \hline
 Baseline&Model&MOTA (\%) &FPS\\
 \hline
 \multirow{2}{*}{Teacher-only}& DLA34 & 68.30 & 20.78 \\
 &YOLOv5& 62.40 & 40.46 \\
 \hline
 \multirow{2}{*}{Student-only}&DLA34&61.40&37.69\\
 &YOLOv5& 38.90 & 59.32 \\
 \hline
\end{tabular}
\label{table:fullmodels}
\end{table}

\begin{table}[h]
\footnotesize
\centering
\caption{AttTrack model experiments with K = 6 on the MOT17 dataset}
 \begin{tabular}{c|c|c|c|c|c|c}
 \hline
 Model&\multicolumn{2}{c}{AttTrack}&\multicolumn{2}{c}{AttTrack-no-update}&\multicolumn{2}{c}{Naive-mix}\\
 \cline{2-7}
  &MOTA&FPS&MOTA&FPS&MOTA&FPS\\
 \hline
 DLA34&63.00&30.80&62.90&31.30&61.40&31.65\\
 YOLOv5&43.60&50.90&43.40&52.24&38.60&53.13\\
 \hline
\end{tabular}
\label{table:attention_or_not}
\end{table}

\begin{table}[h]
\footnotesize
\centering
\caption{YOLOv5 and DLA34 models with IFM on the MOT17 dataset}
 \begin{tabular}{c|c|c|c|c}
 \hline
 $K$ & \multicolumn{2}{c}{YOLOv5-small} & \multicolumn{2}{c}{DLA34-small}\\
 \cline{2-5}
 &MOTA&FPS&MOTA&FPS\\
 \cline{1-5}
 2 & 48.50 &  45.09 & 64.50 & 26.24 \\
 4 & 43.90 & 49.96 & 63.20 & 29.28 \\
 6 & 43.60 & 50.91 & 63.00 & 30.80 \\
 \hline
\end{tabular}
\label{table:fixed_w_size}
\end{table}

\subsection{Alternative teacher}
We conduct further investigations to see whether the representational power of teachers can affect the performance of the student model. Specifically, we compare the use of DLA34 in the teacher model and transfer the knowledge to YOLOv5-small student model. The heatmap computed by a DLA34 teacher can still be useful to the YOLOv5-small student model, and the re-Id features can be aligned using the mechanism in Section \ref{sec:cross_model_feature_learning}.

The DLA34 teacher provides better tracking performance than the YOLOv5-based equivalent, as shown in Table \ref{table:alternative_teacher}, although it runs slower than the YOLOv5 teacher. The gap in tracking performance between YOLOv5 and DLA teacher-based models reduces as $K$ increases as the impact of YOLOv5-small becomes more dominant. Overall, the results in Table \ref{table:alternative_teacher} show that tracking performance and processing time can be considerably impacted by the choice of the teacher architecture. 
\begin{table}[h]
\footnotesize
\centering
\caption{Compression of Different Teacher Models using EFM on the MOT17 dataset.}
 \begin{tabular}{c|c|c|c|c}
 \hline
 \makecell{K}&\multicolumn{2}{c}{YOLOv5 $\rightarrow$YOLOv5-small}&\multicolumn{2}{c}{DLA34 $\rightarrow$YOLOv5-small}\\
 \cline{2-5}
  &MOTA&FPS&MOTA&FPS\\
 \hline
 2& 50.40 & 44.69 & 52.00 & 28.72 \\
 4& 46.90 & 47.60 & 47.70 & 36.41 \\
 6& 45.90 & 48.83 & 46.20 & 40.32 \\
 
 \hline
\end{tabular}
\label{table:alternative_teacher}
\end{table}
\begin{table}[h]
\footnotesize
\centering
\caption{Importance of cross-model feature learning on the MOT17 dataset. EFM: employing an additional convolution layer to translate characteristics from the teacher and student models to the common features space. IFM: student model mimics re-ID feature generated by the student model.}
\resizebox{\columnwidth}{!}{%
 \begin{tabular}{c|c|c|c|c|c|c|c}
 \hline
 Model&\multirow{2}{*}{K}&\multicolumn{2}{c}{EFM}&\multicolumn{2}{c}{IFM}&\multicolumn{2}{c}{No Fea. Learning}\\
 \cline{3-8}
  &&MOTA&FPS&MOTA&FPS&MOTA&FPS\\
 \hline
 \multirow{3}{*}{DLA34}
 &2& 65.30 & 25.40 & 64.50 & 26.24 & 64.30 & 26.30\\
 &4& 64.10 & 28.36 & 63.20 & 29.28 & 63.20 & 29.59\\
 &6& 63.80 & 29.99 & 63.10 & 30.80 & 63.20 & 31.01\\
 \hline
 \multirow{3}{*}{YOLOv5}
 &2& 50.40 & 44.69 & 48.50 & 45.09 & 47.50 & 45.57\\
 &4& 46.90 & 47.60 & 43.90 & 49.96 & 43.20 & 50.22\\
 &6& 45.90 & 48.83 & 43.60 & 50.91 & 42.90 & 51.03\\
 
 \hline
\end{tabular}}
\label{table:cross_model_feature_learning}
\end{table}
\subsection{Cross-model feature learning}
The usefulness of the learned features for transferring knowledge between teacher and student models is evaluated in Table \ref{table:cross_model_feature_learning}. In the experiments, EFM is done by applying a single linear layer on the generated re-ID features. As can be observed, the inclusion of this extra layer reduces the visual feature distance between the teacher and the student, and produces more precise tracking output. The use of EFM for $K=4$ has the greatest influence on the YOLOv5 student model, accounting for 3.7\% of more accurate tracking performance. As we can see, cross-model feature learning is beneficial and has more impacts on YOLOv5 than it does on DLA34-small. Furthermore, EFM yields better tracking performance than IFM but incurs higher computation costs. On DLA34-small, for instance, utilizing EFM with $K=2$ achieves a 65.30\% MOTA score and 25.40 frame rate, whereas the use of IFM results in a 0.9\% lower MOTA score and a 0.84 faster FPS. 

\subsection{Comparison with layer-wise attention transfer}
\begin{figure}[t]
\centering
\subfloat[Tracking performance comparison for DLA34 architecture]{\includegraphics[scale=0.39]{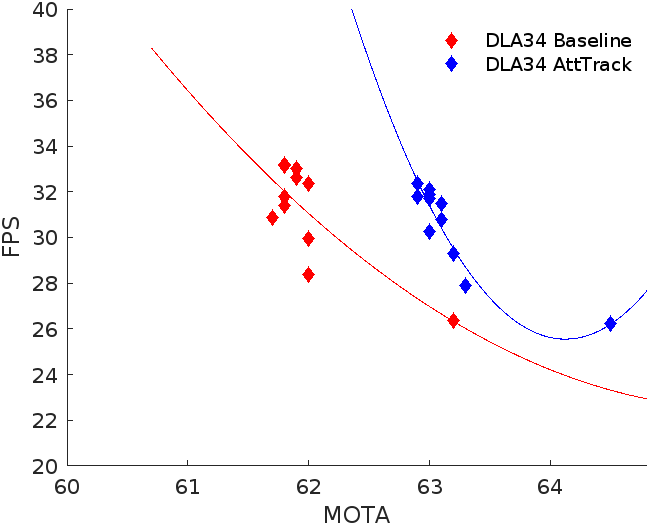}}
\subfloat[Tracking performance comparison for YOLOv5 architecture]{\includegraphics[scale=0.39]{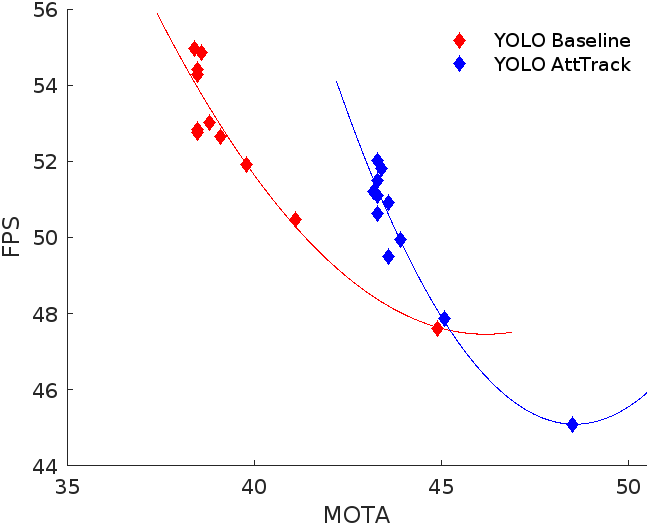}}%
\caption{Results of attention transfer for AttTrack and Layerwise under 11 different $K$s $\in[2,12]$ on the MOT17 dataset}
\label{fig:comp_figure}
\end{figure}

In this set of experiments, we compare AttTrack with the layer-wise attention transfer proposed in~\cite{zagoruyko2016paying} (baseline Layerwise). The main difference between our approach and~\cite{zagoruyko2016paying} is that the layer-wise solution transfers attention knowledge to the student model during the training time only, and the student model performs tracking entirely on its own, while our AttTrack leverages teacher knowledge in both training and inference phases. We implement a layer-wise attention approach for the MOT task since~\cite{zagoruyko2016paying} is originally built for object classification tasks. The results are shown in Figure \ref{fig:comp_figure} for 11 different $K$s between two and twelve. For fair comparison, in the baseline layer-wise attention transfer, we also invoke the teacher model every $K$ frames though there is no knowledge transfer between the teacher and the student models at inference time. 2nd order polynomial fitting functions for the AttTrack and baseline results are also displayed in the figures. When comparing AttTrack to layer-wise attention, we find that AttTrack exceeds the baseline significantly with comparable processing time on tracking accuracy. The difference with YOLOv5 is more pronounced. 
For example, AttTrack is 4 percent better with only 2 percent lower FPS when $K$ is between two and four. The gap in computation time between AttTrack and baseline drops for the DLA34-based tracker. This is because the overhead of attention transfer in AttTrack is shadowed by the high compute cost of the DLA34 backbone. 
\subsection{Experiments on MOT15}
To verify the generalizability of AttTrack to other datasets, we further conduct experiments on MOT15. The performance of Teacher-only and Student-only is given in Table \ref{table:baseline_models_mot15}, and the comparison between AttTrack and Layerwise  for different $K$'s is given in Table \ref{table:mot_15_exps}. Similar to the trends with MOT17, we observe that AttTrack outperforms Layerwise and Student-only in MOTA, and is considerably faster than Teacher-only.

\begin{table}[h]
\footnotesize
\centering
\caption{Performance of Teacher-only and Student-only baselines on the MOT15 dataset}
 \begin{tabular}{c|c|c|c}
 \hline
 Baseline&Model&MOTA (\%) &FPS\\
 \hline
 \multirow{2}{*}{Teacher-only}& DLA34 & 68.80 & 21.70 \\
 &YOLOv5& 61.00 & 54.41 \\
 \hline
 \multirow{2}{*}{Student-only}&DLA34&66.90&41.47\\
 &YOLOv5& 52.70 & 58.80 \\
 \hline
\end{tabular}
\label{table:baseline_models_mot15}
\end{table}

\begin{table}[h]
\footnotesize
\centering
\caption{AttTrack and Layerwise on the MOT15 dataset}
 \begin{tabular}{c|c|c|c|c|c}
 \hline
 Model&\multirow{2}{*}{K}&\multicolumn{2}{c}{AttTrack-EFM}&\multicolumn{2}{c}{Layerwise}\\
 \cline{3-6}
  &&MOTA&FPS&MOTA&FPS\\
 \hline
 \multirow{3}{*}{DLA34}
 &2& 67.60 & 27.99 & 65.90 & 28.64 \\
 &4& 67.40 & 32.32 & 65.70 & 32.55 \\
 &6& 67.30 & 35.19 & 65.60 & 35.90 \\
 \hline
 \multirow{3}{*}{YOLOv5}
 &2& 56.50 & 55.17 & 55.00 & 56.63 \\
 &4& 54.10 & 56.13 & 52.60 & 57.93 \\
 &6& 52.90 & 56.46 & 52.30 & 58.39 \\
  \hline
\end{tabular}
\label{table:mot_15_exps}
\end{table}
\section{Conclusion}
\label{conclusion}
AttTrack is a teacher-student attention transfer approach for accelerating multi-object tracking tasks. It transfers knowledge from a complex teacher to a lightweight student model in both the training and inference stages. AttTrack is model agnostic and can be used in conjunction with other techniques to accelerate neural network inference. Because AttTrack adopts cross-model feature learning, it is capable to transfer knowledge from any teacher to any student network with different network architectures (e.g. YOLOv5 or DLA34). When compared with traditional attention-based methods, our work improves tracking accuracy with marginal degradation in inference time. As part of future work, we are interested in investigating attention mechanisms with adaptive window sizes.

{\small
\bibliographystyle{ieee_fullname}
\bibliography{egbib}

\begin{thebibliography}{10}\itemsep=-1pt

\bibitem{motchallengeweb}
\href{https://motchallenge.net}{{MOT Challenge Website}}.
\newblock https://motchallenge.net.

\bibitem{cai2017deep}
Zhaowei Cai, Xiaodong He, Jian Sun, and Nuno Vasconcelos.
\newblock {Deep Learning With Low Precision by Half-wave Gaussian
  Quantization}.
\newblock In {\em Proceedings of the IEEE conference on computer vision and
  pattern recognition}, pages 5918--5926, 2017.

\bibitem{chen2017learning}
Guobin Chen, Wongun Choi, Xiang Yu, Tony Han, and Manmohan Chandraker.
\newblock Learning efficient object detection models with knowledge
  distillation.
\newblock {\em Advances in neural information processing systems}, 30, 2017.

\bibitem{chen2018reverse}
Shuhan Chen, Xiuli Tan, Ben Wang, and Xuelong Hu.
\newblock Reverse attention for salient object detection.
\newblock In {\em Proceedings of the European conference on computer vision
  (ECCV)}, pages 234--250, 2018.

\bibitem{cheng2018model}
Yu Cheng, Duo Wang, Pan Zhou, and Tao Zhang.
\newblock {Model Compression and Acceleration for Deep Neural Networks: The
  Principles, Progress, and challenges}.
\newblock {\em IEEE Signal Processing Magazine}, 35(1):126--136, 2018.

\bibitem{chin2019adascale}
Ting-Wu Chin, Ruizhou Ding, and Diana Marculescu.
\newblock {Adascale: Towards Real-time Video Object Detection Using Adaptive
  Scaling}.
\newblock {\em arXiv preprint arXiv:1902.02910}, 2019.

\bibitem{dai2021dynamic}
Xiyang Dai, Yinpeng Chen, Jianwei Yang, Pengchuan Zhang, Lu Yuan, and Lei
  Zhang.
\newblock Dynamic detr: End-to-end object detection with dynamic attention.
\newblock In {\em Proceedings of the IEEE/CVF International Conference on
  Computer Vision}, pages 2988--2997, 2021.

\bibitem{denil2012learning}
Misha Denil, Loris Bazzani, Hugo Larochelle, and Nando de Freitas.
\newblock Learning where to attend with deep architectures for image tracking.
\newblock {\em Neural computation}, 24(8):2151--2184, 2012.

\bibitem{ellis2010pets2010}
Anna Ellis and James Ferryman.
\newblock {PETS2010} and {PETS2009} evaluation of results using individual
  ground truthed single views.
\newblock In {\em 2010 7th IEEE international conference on advanced video and
  signal based surveillance}, pages 135--142. IEEE, 2010.

\bibitem{fiaz2020improving}
Mustansar Fiaz, Arif Mahmood, Ki~Yeol Baek, Sehar~Shahzad Farooq, and Soon~Ki
  Jung.
\newblock Improving object tracking by added noise and channel attention.
\newblock {\em Sensors}, 20(13):3780, 2020.

\bibitem{geiger2012we}
Andreas Geiger, Philip Lenz, and Raquel Urtasun.
\newblock Are we ready for autonomous driving? the kitti vision benchmark
  suite.
\newblock In {\em 2012 IEEE conference on computer vision and pattern
  recognition}, pages 3354--3361. IEEE, 2012.

\bibitem{gholami2021survey}
Amir Gholami, Sehoon Kim, Zhen Dong, Zhewei Yao, Michael~W Mahoney, and Kurt
  Keutzer.
\newblock {A survey of Quantization Methods for Efficient neural Network
  Inference}.
\newblock {\em arXiv preprint arXiv:2103.13630}, 2021.

\bibitem{girshick2014rich}
Ross Girshick, Jeff Donahue, Trevor Darrell, and Jitendra Malik.
\newblock {Rich feature hierarchies for accurate object detection and semantic
  segmentation}.
\newblock In {\em Proceedings of the IEEE conference on computer vision and
  pattern recognition}, pages 580--587, 2014.

\bibitem{hinton2015distilling}
Geoffrey Hinton, Oriol Vinyals, and Jeff Dean.
\newblock {Distilling The Knowledge In A Neural Network}.
\newblock {\em arXiv preprint arXiv:1503.02531}, 2015.

\bibitem{huang2021siamatl}
Bo Huang, Tingfa Xu, Ziyi Shen, Shenwang Jiang, Bingqing Zhao, and Ziyang Bian.
\newblock Siamatl: online update of siamese tracking network via attentional
  transfer learning.
\newblock {\em IEEE Transactions on Cybernetics}, 2021.

\bibitem{huang2017like}
Zehao Huang and Naiyan Wang.
\newblock Like what you like: Knowledge distill via neuron selectivity
  transfer.
\newblock {\em arXiv preprint arXiv:1707.01219}, 2017.

\bibitem{hubara2017quantized}
Itay Hubara, Matthieu Courbariaux, Daniel Soudry, Ran El-Yaniv, and Yoshua
  Bengio.
\newblock {Quantized Neural Networks: Training Neural Networks With Low
  Precision Weights and Activations}.
\newblock {\em The Journal of Machine Learning Research}, 18(1):6869--6898,
  2017.

\bibitem{jocher2021ultralytics}
Glenn Jocher, Alex Stoken, Jirka Borovec, Ayush Chaurasia, L Changyu, AV
  Laughing, A Hogan, J Hajek, L Diaconu, YK Marc, et~al.
\newblock ultralytics/yolov5: v5. 0-yolov5-p6 1280 models aws supervise. ly and
  youtube integrations.
\newblock {\em Zenodo}, 11, 2021.

\bibitem{larochelle2010learning}
Hugo Larochelle and Geoffrey~E Hinton.
\newblock Learning to combine foveal glimpses with a third-order boltzmann
  machine.
\newblock {\em Advances in neural information processing systems}, 23, 2010.

\bibitem{li2016pruning}
Hao Li, Asim Kadav, Igor Durdanovic, Hanan Samet, and Hans~Peter Graf.
\newblock {Pruning Filters for Efficient Convnets}.
\newblock {\em arXiv preprint arXiv:1608.08710}, 2016.

\bibitem{li2017mimicking}
Quanquan Li, Shengying Jin, and Junjie Yan.
\newblock Mimicking very efficient network for object detection.
\newblock In {\em Proceedings of the ieee conference on computer vision and
  pattern recognition}, pages 6356--6364, 2017.

\bibitem{liu2020continuous}
Miaomiao Liu, Xianzhong Ding, and Wan Du.
\newblock {Continuous, Real-Time Object Detection on Mobile Devices without
  Offloading}.
\newblock In {\em 2020 IEEE 40th International Conference on Distributed
  Computing Systems (ICDCS)}, pages 976--986. IEEE, 2020.

\bibitem{mehta2018object}
Rakesh Mehta and Cemalettin Ozturk.
\newblock Object detection at 200 frames per second.
\newblock In {\em Proceedings of the European Conference on Computer Vision
  (ECCV) Workshops}, pages 0--0, 2018.

\bibitem{meng2020arnet}
Yue Meng, Chung-Ching Lin, Rameswar Panda, Prasanna Sattigeri, Leonid
  Karlinsky, Aude Oliva, Kate Saenko, and Rogerio Feris.
\newblock {Ar-net: Adaptive Frame Resolution For Efficient Action Recognition}.
\newblock In {\em European Conference on Computer Vision}, pages 86--104.
  Springer, 2020.

\bibitem{molchanov2016pruning}
Pavlo Molchanov, Stephen Tyree, Tero Karras, Timo Aila, and Jan Kautz.
\newblock {Pruning Convolutional Neural Networks for Resource Efficient
  Inference}.
\newblock {\em arXiv preprint arXiv:1611.06440}, 2016.

\bibitem{nalaiedeepscale}
Keivan Nalaie, Renjie Xu, and Rong Zheng.
\newblock {DeepScale: Online Frame Size Adaptation for Multi-object Tracking on
  Smart Cameras and Edge Servers}.

\bibitem{rastegari2016xnor}
Mohammad Rastegari, Vicente Ordonez, Joseph Redmon, and Ali Farhadi.
\newblock {Xnor-net: Imagenet Classification Using Binary Convolutional Neural
  Networks}.
\newblock In {\em European conference on computer vision}, pages 525--542.
  Springer, 2016.

\bibitem{redmon2016you}
Joseph Redmon, Santosh Divvala, Ross Girshick, and Ali Farhadi.
\newblock {You Only Look Once: Unified, Real-time Object Detection}.
\newblock In {\em Proceedings of the IEEE conference on computer vision and
  pattern recognition}, pages 779--788, 2016.

\bibitem{fasterrcnn}
Shaoqing Ren, Kaiming He, Ross Girshick, and Jian Sun.
\newblock {Faster R-CNN: Towards Real-time Object Detection with Region
  Proposal Networks}.
\newblock {\em IEEE transactions on pattern analysis and machine intelligence},
  39(6):1137--1149, 2016.

\bibitem{rensink2000dynamic}
Ronald~A Rensink.
\newblock The dynamic representation of scenes.
\newblock {\em Visual cognition}, 7(1-3):17--42, 2000.

\bibitem{romero2014fitnets}
Adriana Romero, Nicolas Ballas, Samira~Ebrahimi Kahou, Antoine Chassang, Carlo
  Gatta, and Yoshua Bengio.
\newblock Fitnets: Hints for thin deep nets.
\newblock {\em arXiv preprint arXiv:1412.6550}, 2014.

\bibitem{thanikasalam2019target}
Kokul Thanikasalam, Clinton Fookes, Sridha Sridharan, Amirthalingam Ramanan,
  and Amalka Pinidiyaarachchi.
\newblock Target-specific siamese attention network for real-time object
  tracking.
\newblock {\em IEEE Transactions on Information Forensics and Security},
  15:1276--1289, 2019.

\bibitem{wang2019salient}
Wenguan Wang, Shuyang Zhao, Jianbing Shen, Steven~CH Hoi, and Ali Borji.
\newblock {Salient object detection with pyramid attention and salient edges}.
\newblock In {\em Proceedings of the IEEE/CVF Conference on Computer Vision and
  Pattern Recognition}, pages 1448--1457, 2019.

\bibitem{wang2019towards}
Xudong Wang, Zhaowei Cai, Dashan Gao, and Nuno Vasconcelos.
\newblock {Towards universal object detection by domain attention}.
\newblock In {\em Proceedings of the IEEE/CVF Conference on Computer Vision and
  Pattern Recognition}, pages 7289--7298, 2019.

\bibitem{xu2018deepcache}
Mengwei Xu, Mengze Zhu, Yunxin Liu, Felix~Xiaozhu Lin, and Xuanzhe Liu.
\newblock {DeepCache: Principled Cache for Mobile Deep Vision}.
\newblock In {\em Proceedings of the 24th Annual International Conference on
  Mobile Computing and Networking}, pages 129--144, 2018.

\bibitem{ying2019multi}
Xiang Ying, Qiang Wang, Xuewei Li, Mei Yu, Han Jiang, Jie Gao, Zhiqiang Liu,
  and Ruiguo Yu.
\newblock Multi-attention object detection model in remote sensing images based
  on multi-scale.
\newblock {\em IEEE Access}, 7:94508--94519, 2019.

\bibitem{yu2018deep}
Fisher Yu, Dequan Wang, Evan Shelhamer, and Trevor Darrell.
\newblock {Deep Layer Aggregation}.
\newblock In {\em Proceedings of the IEEE conference on computer vision and
  pattern recognition}, pages 2403--2412, 2018.

\bibitem{zagoruyko2016paying}
Sergey Zagoruyko and Nikos Komodakis.
\newblock Paying more attention to attention: Improving the performance of
  convolutional neural networks via attention transfer.
\newblock {\em arXiv preprint arXiv:1612.03928}, 2016.

\bibitem{zhang2020fairmot}
Yifu Zhang, Chunyu Wang, Xinggang Wang, Wenjun Zeng, and Wenyu Liu.
\newblock {FairMOT: On The Fairness Of Detection And Re-Identification In
  Multiple Object Tracking}.
\newblock {\em arXiv preprint arXiv:2004.01888}, 2020.

\bibitem{zhang2021toward}
Zhipeng Zhang, Yufan Liu, Bing Li, Weiming Hu, and Houwen Peng.
\newblock Toward accurate pixelwise object tracking via attention retrieval.
\newblock {\em IEEE Transactions on Image Processing}, 30:8553--8566, 2021.

\bibitem{zhou2019deconstructing}
Hattie Zhou, Janice Lan, Rosanne Liu, and Jason Yosinski.
\newblock {Deconstructing Lottery Tickets: Zeros, Signs, and the Supermask}.
\newblock {\em Advances in neural information processing systems}, 32, 2019.

\end{thebibliography}
}
\end{document}


\title{AttTrack: Online Deep Attention Transfer for Multi-object Tracking\\ Supplementary Material}  

\maketitle

\section{Experiments on HRNet-W18\cite{cheng2020higherhrnet}}
In Table \ref{table:baselines_hrnet}, the results of Teacher-only and Student-only baselines on the MOT15 and MOT17 datasets are reported. The performance gap between large and small models in MOT17 is larger than in the MOT15 dataset. The large model's capability for locating and characterizing small objects in high-resolution images is the primary cause. 
\begin{table}[h]
\footnotesize
\centering
\caption{Performance of Teacher-only and Student-only on MOT15 and MOT17 datasets.}
 \begin{tabular}{c|c|c|c}
 \hline

 Baseline&Dataset&MOTA (\%) &FPS\\
 \hline
 \multirow{2}{*}{Teacher-only}
 &MOT15 & 64.50 & 12.47 \\
 &MOT17& 64.20 & 12.30 \\
 \hline
 \multirow{2}{*}{Student-only}
 & MOT15& 61.30 & 19.99\\
 &MOT17& 55.60 & 19.49 \\
 \hline
\end{tabular}
\label{table:baselines_hrnet}
\end{table}

Table \ref{table:attTrack_hrnet} shows AttTrack's performance with EFM settings. The MOTA gap between $K=2$ and $K=6$ on MOT15 is only 1.60\% while this metric for MOT17 is 3.40\%. We believe these performance differences appear to be related to the corresponding teacher and student baseline models presented in Table \ref{table:baselines_hrnet}. Our attention transfer approach can improve the tracking performance of the student model by efficiently extracting and transforming attention to the student model. We substitute Layerwise and Naive-Mix for AttTrack as two different baselines to assess the effectiveness of our attention transmission approach.  As shown in Table\ref{table:attTrack_hrnet}, AttTrack outperforms Layerwise and Naive-Mix approaches. 

\begin{table}[hbt!]
\footnotesize
\centering
\caption{Comparing AttTrack with Layerwise and Naive-Mix on HRNet-W18 architecture}
\resizebox{\columnwidth}{!}{%
 \begin{tabular}{c|c|c|c|c|c|c|c}
 \hline
 Dataset&\multirow{2}{*}{K}&\multicolumn{2}{c}{AttTrack-EFM}&\multicolumn{2}{c}{Layerwise}&\multicolumn{2}{c}{Naive-Mix}\\
 \cline{3-8}
  &&MOTA&FPS&MOTA&FPS&MOTA&FPS\\
  \hline
 \multirow{5}{*}{MOT15}
 &2& \textbf{63.50} & 15.58 & 62.70 & 15.47 & 62.60 & 15.59\\
 &3& 62.50 & 16.58 & 61.50 & 16.89 & 61.60 & 17.00\\
 &4& 62.10 & 17.64 & 61.50 & 17.56 & 61.10 & 17.77\\
 &5& 62.00 & 17.88 & 61.60 & 18.00 & 60.90 & 18.17\\
 &6& 61.90 & 18.04 & 61.30 & 18.50 & 61.10 & 18.52\\
  \hline
\multirow{5}{*}{MOT17}
 &2& \textbf{60.10} & 14.84 & 59.90 & 15.14 & 58.90 & 15.16\\
 &3& 58.10 & 15.93 & 57.30 & 16.28 & 56.20 & 16.30\\
 &4& 57.50 & 16.65 & 56.90 & 17.09 & 56.10 & 17.00\\
 &5& 56.90 & 17.16 & 56.00 & 17.51 & 55.70 & 17.36\\
 &6& 56.70 & 17.40 & 56.70 & 17.91 & 55.80 & 17.50\\
  \hline
\end{tabular}}
\label{table:attTrack_hrnet}
\end{table}

{\small
\bibliographystyle{ieee}
\bibliography{egbibrebuttal}
}